\documentclass[11pt,journal,compsoc]{IEEEtran}
\ifCLASSOPTIONcompsoc
  \usepackage[nocompress]{cite}
\else
  \usepackage{cite}
\fi
\ifCLASSINFOpdf
  \usepackage[pdftex]{graphicx}
\else
\fi
\begin{document}
\title{Domain Adaptation for Ear Recognition Using Deep Convolutional Neural Networks}
\author{\IEEEauthorblockN{Fevziye Irem Eyiokur\footnotemark\textsuperscript{*}, Dogucan Yaman\textsuperscript{*}, Haz{\i}m Kemal Ekenel} 

\IEEEauthorblockA{Department of Computer Engineering \\
Istanbul Technical University, \\
Email: \{eyiokur16, yamand16, ekenel\}@itu.edu.tr} }

\IEEEcompsoctitleabstractindextext{%

\begin{abstract}
In this paper, we have extensively investigated the unconstrained ear recognition problem. We have first shown the importance of domain adaptation, when deep convolutional neural network models are used for ear recognition. To enable domain adaptation, we have collected a new ear dataset using the Multi-PIE face dataset, which we named as Multi-PIE ear dataset. To improve the performance further, we have combined different deep convolutional neural network models. We have analyzed in depth the effect of ear image quality, for example illumination and aspect ratio, on the classification performance. Finally, we have addressed the problem of dataset bias in the ear recognition field. Experiments on the UERC dataset have shown that domain adaptation leads to a significant performance improvement. For example, when VGG-16 model is used and the domain adaptation is applied, an absolute increase of around 10\% has been achieved. Combining different deep convolutional neural network models has further improved the accuracy by 4\%. It has also been observed that image quality has an influence on the results. In the experiments that we have conducted to examine the dataset bias, given an ear image, we were able to classify the dataset that it has come from with 99.71\% accuracy, which indicates a strong bias among the ear recognition datasets.
\end{abstract}
\begin{IEEEkeywords}
Ear recognition, deep learning, domain adaptation
\end{IEEEkeywords}}

\maketitle

\renewcommand{\thefootnote}{\Alph{footnote}}
\footnotetext{\noindent\rule{8cm}{0.4pt}}
\footnotetext{\textsuperscript{*}\textit{The authors have equally contributed.}}
\footnotetext{\textit{This paper is a postprint of a paper submitted to and accepted for publication in IET Biometrics and is subject to Institution of Engineering and Technology Copyright. The copy of record is available at the IET Digital Library.}}

\IEEEdisplaynotcompsoctitleabstractindextext

\IEEEpeerreviewmaketitle

\section{Introduction}\label{sec1}
Human identification through biometrics has been both an important and popular research field. 
Among the biometric traits, ear is a unique part of the human body in terms of different features such as shape, appearance, posture, and there is usually not much change in the ear structure except that the ear length is prolonged over time \cite{Emersic_2017_a}. Various studies have been conducted and many different approaches have been proposed on ear recognition, however, it still remains as an open challenge, especially when the ear images are collected under uncontrolled conditions as in the \textit{Unconstrained Ear Recognition Challenge (UERC)} \cite{Emersic_2017_b}.

Ear recognition approaches are mainly categorized into four groups, \textit{holistic, local, geometric, and hybrid processing} \cite{Emersic_2017_a}. In the earlier studies, the most popular feature extraction methods for ear recognition were SIFT \cite{Hurley_2005}, SURF \cite{Prakash_2013}, and LBP \cite{Wang_2011}. Due to the popularity of deep learning in recent years and its significant impact on the computer vision field \cite{Krizhevsky_2012,Simonyan_2014,Szegedy_2015,Iandola_2016,Galdamez_2016}, deep convolutional neural networks (CNN) based approaches have also been adopted for ear recognition \cite{Emersic_2017_b,Emersic_2017_c,Galdamez_2016}. 
CNNs mainly require a large amount of data for training. However, the amount of samples in the datasets available for ear recognition are rather limited \cite{Emersic_2017_a,Emersic_2017_b,Carreira_Perpinan_1995,Kumar_2012,Frejlichowski_2010,Gonzalez_2008}. 
Due to this limitation, CNN-based ear recognition approaches mainly utilize an already trained object classification model, so called a pretrained deep CNN model, from one of the well-known, high performing CNN architectures, for example \cite{Krizhevsky_2012,Simonyan_2014,Szegedy_2015}. These pretrained models were trained on the ImageNet dataset \cite{Deng_2009}
for generic object classification purposes, therefore, they are required to be adapted to the ear recognition problem. This adaptation is done mainly with a fine-tuning process, where the output classes are updated with subject identities and the employed pretrained deep CNN model is further trained using the training part of an ear dataset.

\begin{figure}[t]
	\centering
	\includegraphics[scale=0.6]{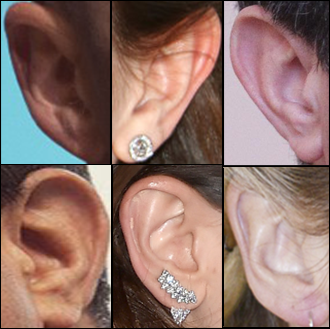}
	\caption{Sample ear images from the UERC dataset \cite{Emersic_2017_b}. The dataset contains many appearance variations in terms of ear direction (left or right), accessories, view angle, image resolution, and illumination.}
	\label{Fig1}
\end{figure}

In the field of ear recognition, most of the used datasets have been collected under controlled conditions, and therefore, very high recognition performance has been achieved on them \cite{Emersic_2017_a}. But the proximity of these accuracies to the real world is a topic of debate. Because of this, \textit{in-the-wild} datasets have been collected in order to imitate real-world challenges confronted in ear recognition better~\cite{Emersic_2017_a}. These datasets, since they contain images collected from the web, have a large variety, for example, in terms of resolution, illumination, and use of accessories. Sample ear images shown in Fig. \ref{Fig1} are from the UERC dataset. It can be seen from Fig. \ref{Fig1} that there are accessories, partial occlusions due to hair, and also pose and illumination variations. Because of these significant appearance variations, the performance of the ear recognition systems on the \textit{wild} datasets, such as on the UERC, is not as high as the ones obtained on the datasets collected under controlled conditions.  

In this paper, we present a comprehensive study on ear recognition \textit{in the wild}. We have employed well-known, high performing deep CNN models, namely, AlexNet \cite{Krizhevsky_2012}, VGG-16 \cite{Simonyan_2014}, and GoogLeNet~\cite{Szegedy_2015} and proposed a domain adaptation strategy for deep CNN-based ear recognition. We have also provided an in depth analysis of several aspects of ear recognition. Our contributions are summarized as follows: 
\begin{itemize}
\item We have proposed a two-stage fine-tuning strategy for domain adaptation. 
\item We have prepared an ear image dataset from Multi-PIE face dataset, which we named as Multi-PIE ear dataset. As can be seen in Table \ref{tab_dataset}, this database contains a larger number of ear images compared to the other ear datasets.
\item We have analyzed the effect of data augmentation and alignment on the ear recognition performance.
\item We have performed deep CNN model combination to improve accuracy.
\item We have examined varying aspect ratios of ear images and the illumination conditions they contain, and assess their influence on the performance.
\item We have investigated the dataset bias problem for ear recognition.

\end{itemize}

For the experiments, we have used the Multi-PIE ear and the UERC datasets \cite{Emersic_2017_b}. Since Multi-PIE ear dataset is collected under controlled conditions, the achieved results were very high. From the experiments on the UERC dataset, we have shown that the proposed two-stage fine-tuning scheme is very beneficial for ear recognition. With data augmentation and without alignment, for AlexNet \cite{Krizhevsky_2012}, the correct classification rate is increased from 52\% to 56.46\%. For VGG-16 \cite{Simonyan_2014} and GoogLeNet \cite{Szegedy_2015}, the increase is from 54.2\% to 63.62\% and from 55.02\% to 60.91\%, respectively. Combining different deep convolutional neural network models has led to further improvement in performance by 4\% compared to the single best performing model. We have observed that data augmentation enhances the accuracy, whereas performing alignment did not improve the performance. However, this point requires further investigation, since only a coarse alignment has been performed by flipping the ear images to one side. Experimental results show that the ear recognition system performs better, when the ear images are cropped from profile faces. Very dark and very bright illumination causes missing details and reflections, which results in performance deteriorations. Experiments to examine the dataset bias have indicated a strong bias among the ear recognition datasets. 

The remainder of the paper is organised as follows. A brief review of the related work on ear recognition is given in Section 2. The employed methods in this work are explained in Section 3. In Section 4, experimental results are presented and discussed. Finally, Section 5 provides conclusions and future research directions.

\section{Related Work}
Many studies have been conducted in the field of ear recognition. In the following paragraphs, we give a brief overview. A comprehensive analysis of the existing studies in the area of ear recognition has been presented in \cite{Emersic_2017_a}. Please refer to this paper for an extensive survey.

In \cite{Emersic_2017_a}, an \textit{in-the-wild} ear recognition dataset AWE and an ear recognition toolbox for MATLAB are introduced. The AWE dataset has become a useful dataset for the ear recognition field, which has previously employed ear datasets that have been collected under controlled conditions. The presented toolbox enables feature extraction from images with traditional, hand-crafted feature extraction methods. The toolbox also provides use of different distance metrics and tools for classification and performance assessment. 

Recently, a competition, \textit{unconstrained ear recognition challenge (UERC)}, was organized \cite{Emersic_2017_b}. The UERC dataset is introduced for this competition. For the benchmark, training and testing sets from this dataset are specified. In the competition, mainly hand-crafted feature extraction methods, such as LBP \cite{Wang_2011} and POEM \cite{Vu_2010}, and CNN-based feature extraction methods are used. One of the proposed methods in this challenge eliminates earrings, hair, other obstacles, and background from the ear image with a binary ear mask. Recognition is performed using the hand-crafted features. In another proposed approach, the score matrices calculated from the CNN-based features and hand-crafted features are fused. The remaining approaches participating to the competition employ only CNN-based features.

In \cite{Guo_2008}, a new feature extraction method named Local Similarity Binary Pattern (LSBP) is introduced. This new method, which is used in conjunction with the Local Binary Pattern (LBP) features, is found to have superior ear recognition performance \cite{Guo_2008}. The proposed feature extraction method provides information both about connectivity and similarity.

In a recent study \cite{Galdamez_2016}, a brief review of deep learning based ear recognition approaches is given. When performed on the ear datasets that contain ear images collected under controlled conditions, deep learning-based approaches provide satisfactory results. However, it has been emphasized that the detection of an ear in the image is a difficult task. 

Another study that employed deep CNN models is presented in~\cite{Emersic_2017_c}. In this work, AlexNet \cite{Krizhevsky_2012}, VGG-16 \cite{Simonyan_2014}, and SqueezeNet \cite{Iandola_2016} architectures are used. Two different training approaches are applied, namely training of the whole model, called full model learning and training of the last layers by using a pretrained deep CNN model, called selective model learning. The best results are obtained with the SqueezeNet. Data augmentation has been applied to increase the amount of data for deep CNN model training. So called selective model learning, using the pretrained models that were trained on the ImageNet dataset, was found to perform better than using so called full model learning in terms of ear recognition performance.

\section{Methodology}\label{sec2}
In this section, we present the employed deep convolutional neural network models, data augmentation and transfer learning approaches, and provide information about the datasets, data alignment, and fusion techniques.

\subsection{Convolutional Neural Networks}\label{subsec2.2}

\begin{figure*}[t]
	\centering{\includegraphics[scale=0.29]{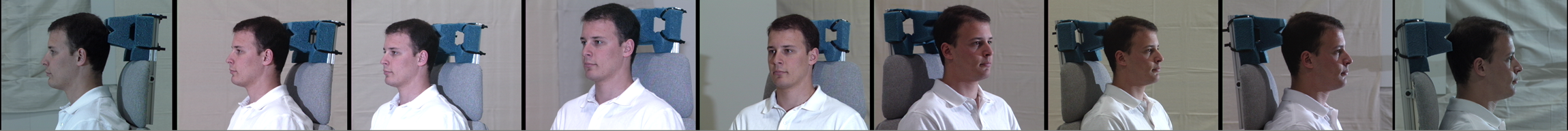}}
	\caption{Selected view angles from the Multi-PIE face dataset \cite{Gross_2008_a, Gross_2008_b}}
	\label{Fig_multi_pie}
\end{figure*}

\begin{figure}[t]
	\centering{\includegraphics[scale=0.38]{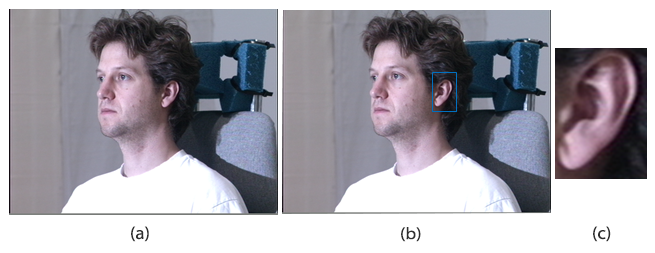}}
	\caption{Illustration of ear detection and  cropping on the Multi-PIE face dataset \cite{Gross_2008_a, Gross_2008_b}: (a) Input image, (b) Ear detected image, (c) Cropped ear image}
	\label{fig_multi_crop}
\end{figure}

In our study, we have employed convolutional neural networks for ear image representation and classification. CNN contains several layers that perform convolution, feature representation, and classification. Convolutional part of the CNNs includes layers that perform many operations, such as convolution, pooling, batch normalization~\cite{Ioffe_2015}, and these layers are sequentially placed to learn the discriminative features from the image. Then, in the later layers, these features are utilized for classification. In this work, for the final layer, we have used the softmax loss in the employed deep CNN models.

The first deep convolutional neural network architecture used in this study is AlexNet \cite{Krizhevsky_2012}, which is the winner model of ILSVRC 2012 challenge \cite{Russakovsky_2015}. In AlexNet \cite{Krizhevsky_2012}, there are five convolutional layers and three fully connected layers. Dropout method \cite{Srivastava_2014} has been used to prevent overfitting. Besides, we have also utilized VGG~\cite{Simonyan_2014} and GoogLeNet \cite{Szegedy_2015} architectures. GoogLeNet \cite{Szegedy_2015} has 22 layers, however, has about twelve times fewer parameters than AlexNet \cite{Krizhevsky_2012}, and it is based on a new paradigm, which is named as inception. In inception layers, input image is filtered by different filters separately. Results of all different filters are utilized, which is very beneficial in terms of extracting multiple features from the same input data. VGG architecture has two versions. One of them contains 16 layers and is named as VGG-16, whereas the other one has 19 layers and is named as VGG-19. VGG-16 has two fully connected layers and softmax classifier after convolutional layers as in AlexNet \cite{Krizhevsky_2012}. VGG-16 \cite{Simonyan_2014} is a deeper network than AlexNet \cite{Krizhevsky_2012} and uses a large number of filters of small size, i.e. $3 \times 3$. 

\subsection{Transfer Learning, Domain Adaptation and Alignment}\label{subsec2.6}
Transfer learning has been applied mainly in two different ways in convolutional neural networks and depends on the size and similarity between the pretraining dataset and the target dataset. The first common approach is to utilize a pretrained deep CNN model directly to extract features from the input images. These extracted features are then fed into, for example, a support vector machine classifier, to learn to discriminate different classes from each other. This scheme is employed when the target dataset contains a small amount of samples. The second approach is fine-tuning the pretrained deep CNN models on the target dataset. That is, to initialize the network weights with the pretrained model and to further train and fine-tune the weights on the target dataset. This method is useful when the target dataset has sufficient amount of training samples, since performing fine-tuning on a target dataset with few training samples can lead to overfitting \cite{Yosinski_2014}. Depending on the task similarity between the two datasets and amount of available training samples in the target dataset, one can decide between these two approaches \cite{Lecun_2015}.

In our work, by using the pretrained models of AlexNet \cite{Krizhevsky_2012}, VGG-16~\cite{Simonyan_2014}, and GoogLeNet \cite{Szegedy_2015} architectures, which were trained on the ImageNet dataset \cite{Deng_2009}, we have fine-tuned them on the ear datasets. The ear recognition datasets contain a limited amount of training samples, for example the ones used in this study contain around a thousand to ten thousand ear images. This amount of training data is sufficient for fine-tuning, although it would not be enough to train a deep CNN model from scratch. In our previous work on age and gender classification \cite{Ozbulak_2016}, we have shown that transferring a pretrained deep CNN model can provide better classification performance than training a task specific CNN model from scratch, when only a limited amount of data is available for the task at hand, as in the case for ear recognition. We have further shown that transferring a CNN model from a closer domain, that is for age and gender classification transferring a pretrained model that were trained on face images, instead of one trained on generic object images, provides better performance. By utilizing this information, we have performed a two-stage fine-tuning of the pretrained deep CNN models for ear recognition. For this approach, we have first constructed an ear dataset from the Multi-PIE face dataset~\cite{Gross_2008_a,Gross_2008_b}. Then, we have fine-tuned the pretrained deep CNN models on this dataset. This way, we first provide a domain adaptation for the pretrained deep CNN models. In the second stage, we perform the final fine-tuning operation by using the target dataset, which is the UERC \cite{Emersic_2017_b} dataset, in this work. This final fine-tuning stage provides a more specific domain and/or task adaptation. In our case, it is the adaptation required for the \textit{wild, uncontrolled} conditions. This step is indeed also very important, since as we have shown in the experiments, there exists a dataset bias \cite{Torralba_2011} among the ear recognition datasets. 

While performing fine-tuning, parameters have been initialized with the values that came from the pretrained network models. The learning rate of last fully connected layer has been increased by ten times. This is a commonly used strategy in fine-tuning, since the early layers mainly focus on low-level feature extraction and the later layers are mainly responsible for classification. 
Global learning rate is selected as 0.0001 for AlexNet \cite{Krizhevsky_2012} and GoogLeNet \cite{Szegedy_2015}, and 0.001 for VGG-16 \cite{Simonyan_2014} during fine-tuning on the Multi-PIE ear and UERC datasets \cite{Emersic_2017_b}. The learning rate is divided by ten in every 20k iterations in AlexNet \cite{Krizhevsky_2012} and VGG-16 \cite{Simonyan_2014}.

Since alignment is a critical factor in visual recognition tasks, to investigate its impact, we have performed fine-tuning with two different setups. In the first one, both right and left side of ear images have been used directly. In the second approach, the training data have been aligned to the same direction and then fine-tuning has been done with these flipped images. That is, all ear images are aligned only to the left side ear or to the right side ear. This setup has been used to reduce the amount of appearance variations within the classes.

\subsection{Data Augmentation}\label{subsec2.3}

Since the number of images in the UERC dataset \cite{Emersic_2017_b} is limited, in order to increase the amount of data as well as to account for appearance variations due to image transformations, we have applied data augmentation.
Data augmentation has also been applied to the Multi-PIE ear dataset. Although the Multi-PIE ear dataset contains around eight times more images than the UERC dataset \cite{Emersic_2017_b}, it would still benefit from data augmentation. In this work, data augmentation is performed by using the Imgaug tool\footnote{http://github.com/aleju/imgaug}. 

For data augmentation, different transformations have been used and many images have been created from a single image. First of all, some images that are $224 \times 224$ pixel resolution are randomly cropped from images of size $256 \times 256$ pixels. Then, in the setup used without alignment, the flipped versions of the images have been produced. Images have been generated at different brightness levels by adding or subtracting values to the pixels' intensity values. These values have been prepared by incrementing by ten in the range of [-55 +55], e.g. (-55, -45, ... +45, +55). Another way of modifying brightness levels of the images have been performed by multiplying the pixels' intensity values with a constant. For this, the values are increased by step size of 0.1 between 0.5 and 1.5. To apply Gaussian blur, we have used different sigma values, which are 0.25, 0.5, 0.75, 1, 1.25, 1.5, 1.75, and 2. 
Sharpening is applied on each image by selecting values from 0.5 to 2.0 with increasing by step size of 0.1 (0.5, 0.6, 0.7 and etc.). This parameter adjusts the lightness/brightness of the output image.
With pixel dropout, from images 
some pixels are dropped and noisy images are created to increase the generalization of the deep learning model. With contrast normalization, images are created in different contrasts. Scale, translate, rotate, and shear methods have been used to increase image variety. For rotation, the angle values in the range of -20 to +20 degrees are used with step size of five degrees. For shear values, again with step size of five degrees, values between -15 and +15 degrees are used. These augmentation parameters are the ones that we have applied to the UERC dataset \cite{Emersic_2017_b}. In the augmentation, we have applied to the Multi-PIE ear dataset, fewer parameters were used. After these processes, roughly 220.000 training images for UERC dataset \cite{Emersic_2017_b} and around 400.000 training images for Multi-PIE ear dataset have been obtained.

\begin{figure*}[t]
	\centering{\includegraphics[scale=0.32]{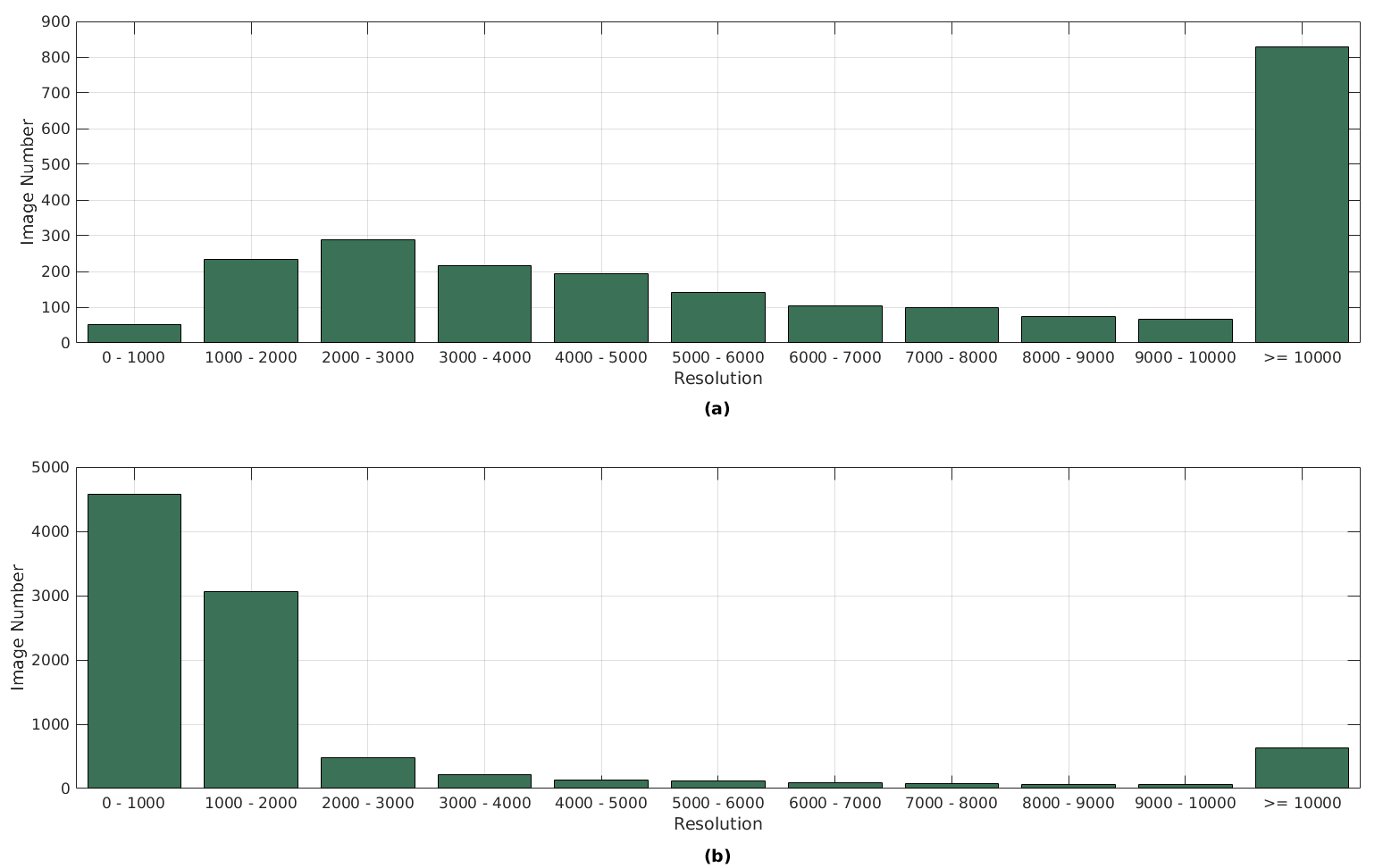}}
	\caption{UERC dataset distribution of number of images with respect to image resolution in (a) training set, (b) test set.}
	\label{fig_UERC_resolution}
\end{figure*}

\begin{figure}[t]
	\centering{\includegraphics[scale=0.40]{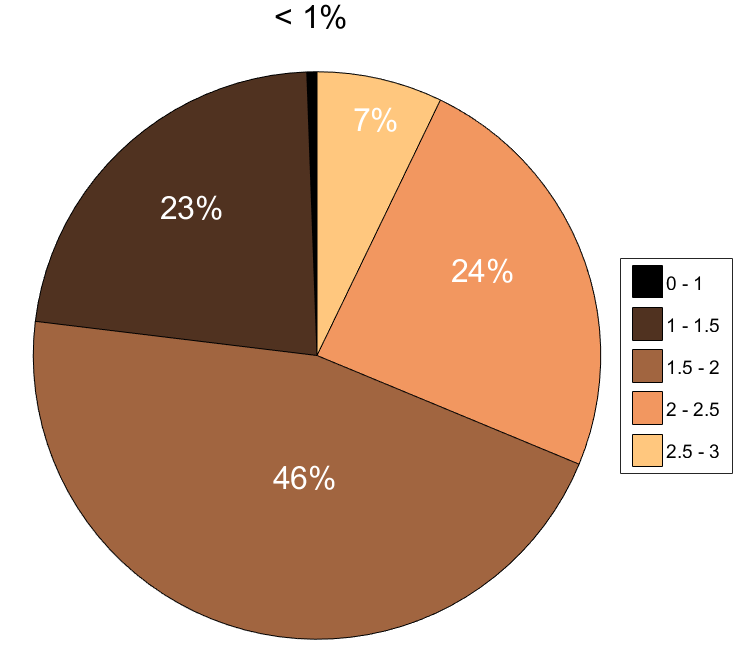}}
	\caption{UERC dataset percentages of ear images with respect to aspect ratio.}
	\label{fig_aspect_ratio}
\end{figure}

\subsection{Datasets}\label{subsec2.1}

\subsubsection{Multi-PIE Ear Dataset} 

Multi-PIE face dataset contains 337 subjects, whose images are acquired, as the name implies, under different pose, illumination, and expression conditions \cite{Gross_2008_a,Gross_2008_b}. Due to the large amount of profile and close-to-profile images available in the Multi-PIE dataset, we have utilized it to create an ear dataset, which we named as Multi-PIE ear dataset\footnote{The list of image filenames and corresponding ear bounding boxes are available at https://github.com/irmdgcn/ear\_recognition}. The view angles that have been selected for ear dataset creation can be seen from Fig. \ref{Fig_multi_pie}. 
Ear detection has been performed using an ear detection implementation for OpenCV \cite{Bradski_2000}. 
A sample ear detection output is shown in Fig. \ref{fig_multi_crop}. 
Since we have used a generic ear detector, the detection accuracy on the Multi-PIE dataset is not very high, 28.3\%, therefore, the ears have been detected successfully only in a subset of the images. Consequently, the new ear dataset that we have obtained from the Multi-PIE face dataset \cite{Gross_2008_a,Gross_2008_b} contains around 17.000 ear images of 205 subjects. This ear dataset has been used for domain adaptation for ear recognition.

\subsubsection{UERC Dataset} 
In the ear recognition field, most of the datasets have been collected under controlled conditions, such as in a laboratory environment. Unlike these datasets, the UERC dataset \cite{Emersic_2017_b} has been collected from the \textit{wild}, that is, it consists of ear images of varying quality collected from the web. Because of this, ear identification problem on the UERC dataset \cite{Emersic_2017_b} is a more challenging task. The UERC dataset is divided into two parts as training and testing sets. In total, there are 11804 ear images of 3706 subjects. Training part of the UERC dataset contains 2304 images of 166 subjects and testing part has 9500 images of 3540 subjects. Following the experimental setup in Emersic et. al. \cite{Emersic_2017_c}, just the training part of this dataset has been used for the experiments. Our experimental results for the test part of the UERC dataset can be found in the unconstrained ear recognition challenge summary paper \cite{Emersic_2017_b}. 
Briefly, we have proposed two approaches in \cite{Emersic_2017_b}. The first one was a CNN-based approach utilizing the VGG-16 architecture, which attained 6.1\% rank-1 recognition rate. The second one, which achieved the best score in our experiments with 6.9\% rank-1 recognition rate, was a fusion-based approach combining the scores from the VGG-16 framework with the ones from the hand-crafted LBP descriptors.
The reason to follow the experimental setup in \cite{Emersic_2017_c}, instead of the one in \cite{Emersic_2017_b}, is the high number of very low resolution images that exist in the test set, which causes problems to interpret the results and analyze the impact of the experimented factors. Distributions of the number of samples with respect to image resolution ---in terms of the total amount of pixels contained in the image--- are given in Fig. \ref{fig_UERC_resolution} for the UERC training and testing datasets separately. As can be seen, most of the ear images in the testing set of the UERC dataset are of low resolution, with the majority containing less than one thousand pixels. UERC training set has a more even distribution and contains more ear images with better resolutions, i.e. having more than ten thousand pixels. The training part of the UERC dataset is created by combining the AWED (1000 images), CVLED (804 images) datasets, and 500 extra images that have been collected from the web \cite{Emersic_2017_a,Emersic_2017_c}. In the rest of the paper, UERC experiments refer to the experiments conducted on the training part of the UERC dataset as in \cite{Emersic_2017_c}. We have also analyzed the aspect ratio vs. number of images in the training part of the UERC dataset. As can be seen in Fig. \ref{fig_aspect_ratio}, aspect ratio of the images varies significantly, due to differences in ear shapes and viewing angles, which makes the unconstrained ear recognition problem even more challenging.

\begin{table}[!t]
\renewcommand{\arraystretch}{1.3}
\caption{Ear datasets}
\label{tab_dataset}
\centering
\begin{tabular}{|c|c|c|}
\hline
Dataset  & \# Images & \# Subjects\\
\hline
AWE \cite{Emersic_2017_a} & 1000 & 100 \\
\hline
AMI \cite{Gonzalez_2008} & 700 & 100 \\
\hline
WPUT \cite{Frejlichowski_2010} & 2071 & 501 \\
\hline
IITD \cite{Kumar_2012} & 493 & 125 \\
\hline
CP \cite{Carreira_Perpinan_1995} & 102 & 17 \\
\hline
UERC Train \cite{Emersic_2017_b} & 2304 & 166 \\
\hline
Multi-PIE Ear \cite{Gross_2008_a,Gross_2008_b} & 17183 & 205 \\
\hline
\end{tabular}
\end{table}

\subsubsection{Other Ear Datasets} 
There are many other ear datasets, which have been collected under controlled conditions, such as Carreira-Perpinan (CP) \cite{Carreira_Perpinan_1995}, Indian Institute of Technology Delhi (IITD) \cite{Kumar_2012}, AMI \cite{Gonzalez_2008}, West Pommeranian University of Technology (WPUT) \cite{Frejlichowski_2010}, and AWE \cite{Emersic_2017_a} datasets as listed in Table \ref{tab_dataset}. The CP dataset \cite{Carreira_Perpinan_1995} contains 102 images belonging to 17 subjects. All ear images in this dataset have been captured from the left side. There exists no accessories or occlusions. The second dataset IITD \cite{Kumar_2012} contains 493 images of 125 subjects and all images are from the right side of the ear. Accessories exist in this dataset. The third one is AMI \cite{Gonzalez_2008} and contains 700 images of 100 different subjects. Both sides of the ears are available in this dataset. However, there are no accessories. Another ear dataset is WPUT \cite{Frejlichowski_2010} that includes 501 subjects and 2071 ear images. Accessories exist in this dataset. The last one is AWE dataset \cite{Emersic_2017_a}, which is also included in the UERC dataset. There are 1000 ear images of 100 subjects in this dataset. These 100 subjects are the first 100 subjects of the UERC dataset \cite{Emersic_2017_b,Emersic_2017_c}. Many studies have been conducted on these datasets in previous studies and the performance of the proposed approaches on the ones that are collected under controlled conditions are very high. However, as shown in the \textit{unconstrained ear recognition challenge} \cite{Emersic_2017_b}, \textit{ear recognition in-the-wild} poses several difficulties causing lower recognition accuracies. In our work, along with the Multi-PIE ear dataset and the UERC dataset, we have utilized these other datasets, especially to investigate whether there exists a dataset bias in the ear recognition field. Sample images from these datasets can be seen in Fig. \ref{fig_all_dataset}.

\begin{table}[!b]
\renewcommand{\arraystretch}{1.3}
\caption{Confidence score calculation formulas}
\label{tab_fusion_method}
\centering
\begin{tabular}{|c|c|}
\hline
Name  & Formula \\
\hline
Basic & $c = s[0]$ \\
\hline
d2s & $c = s[0] - s[1]$ \\
\hline
d2sr & $c = 1 - (s[1]/s[0])$ \\
\hline
avg-diff & $c = \frac{1}{M-1} \sum_{i=1}^M (s[0] - s[i])$ \\
\hline
diff1 & $c = \sum_{i=1}^{M-1} (\frac{s[i-1] - s[i]}{i})$ \\
\hline
\end{tabular}
\end{table}

\subsection{Fusion}\label{subsec2.8}

In order to improve the accuracy further, we have utilized model fusion. The classification outputs of different deep CNN models are combined according to their confidence scores for each image. We have employed different confidence score calculation methods as listed in Table \ref{tab_fusion_method}. In the table, array \textit{s} contains prediction percentages obtained by the model in a sorted order ---large to small---, that is, it contains \textit{raw} classification scores. The array \textit{c} contains confidence scores, which are calculated by using the formulas listed in the table. The deep CNN model with the highest confidence score for an image is accepted as the most reliable model for that image. 
In this work, model combination is applied for the experiments on the UERC dataset \cite{Emersic_2017_b}. AlexNet \cite{Krizhevsky_2012}, VGG-16 \cite{Simonyan_2014}, and GoogLeNet \cite{Szegedy_2015} models are combined with each other.

\begin{table}[t]
\renewcommand{\arraystretch}{1.3}
\caption{Multi-PIE ear dataset test results}
\label{Multi-Pie}
\centering
\begin{tabular}{|c|c|c|c|}
\hline
Models & Accuracy & Augmentation & Alignment\\
\hline
AlexNet & 96.71\% & + & + \\
\hline
AlexNet & 99.81\% & + & $\times$ \\
\hline
AlexNet & 97.64\% & $\times$ & $\times$ \\
\hline
VGG-16 & 100\% & + & + \\
\hline
VGG-16 & 100\% & + & $\times$ \\
\hline
VGG-16 & 98.57\% & $\times$ & $\times$ \\
\hline
GoogLeNet & 97.80\% & + & + \\
\hline
GoogLeNet & 99.32\% & + & $\times$ \\
\hline
GoogLeNet & 98.45\% & $\times$ & $\times$ \\
\hline
\end{tabular}
\end{table}

\begin{table}[!b]
\renewcommand{\arraystretch}{1.3}
\caption{UERC dataset test results}
\label{UERC_results}
\centering
\begin{tabular}{|c|c|c|c|c|}
\hline
Models & Accuracy & Fine-Tuning & Aug. & Align\\
\hline
AlexNet \cite{Emersic_2017_c} & 49.51\% & ImageNet & + & $\times$ \\
\hline
VGG-16 \cite{Emersic_2017_c} & 51.25\% & ImageNet & + & $\times$ \\
\hline
SqueezeNet \cite{Emersic_2017_c} & 62.00\% & ImageNet & + & $\times$ \\
\hline
AlexNet & 49.51\% & ImageNet & $\times$ & $\times$ \\
\hline
AlexNet & 52.00\%  & ImageNet & + & $\times$ \\
\hline
AlexNet & 53.20\% & Multi-PIE & $\times$ & $\times$ \\
\hline
AlexNet & 56.46\% & Multi-PIE & + & $\times$ \\
\hline
AlexNet & 56.02\% & Multi-PIE & + & + \\
\hline
VGG-16 & 51.03\% & ImageNet & $\times$ & $\times$ \\
\hline
VGG-16 & 54.2\% & ImageNet & + & $\times$ \\
\hline
VGG-16 & 58.84\% & Multi-PIE & $\times$ & $\times$ \\
\hline
VGG-16 & 63.62\% & Multi-PIE & + & $\times$\\
\hline
VGG-16 & 62.64\% & Multi-PIE & + & + \\
\hline
GoogLeNet & 54.72\% & ImageNet & $\times$ & $\times$ \\
\hline
GoogLeNet & 55.02\% & ImageNet & + & $\times$ \\
\hline
GoogLeNet & 55.37\% & Multi-PIE & $\times$ & $\times$ \\
\hline
GoogLeNet & 60.91\% & Multi-PIE & + & $\times$ \\
\hline
GoogLeNet & 60.58\% & Multi-PIE & + & + \\
\hline
\end{tabular}
\end{table}

\section{Experimental Results}\label{sec3}
We have conducted the ear recognition experiments on the Multi-PIE ear dataset and the UERC dataset \cite{Emersic_2017_b}. The other ear datasets have been used to assess dataset bias. Multi-PIE ear dataset is divided into three parts as train, validation, and test set. 80\% of the dataset has been used for training, 10\% has been used for validation, and the remaining 10\% has been employed for testing. The experimental setup for the experiments on the UERC dataset \cite{Emersic_2017_b} is the same as the one in Emersic et. al. \cite{Emersic_2017_c}. $60\%$ of the dataset has been used for training and the remaining $40\%$ has been used for testing. Data augmentation and alignment have been applied on the training part of the Multi-PIE ear dataset and the UERC dataset \cite{Emersic_2017_b}.

In the experiments, for deep convolutional neural network model training, images have been resized to $256 \times 256$ pixels resolution. These $256 \times 256$ sized images are cropped into five different images during the training phase and a single crop is taken from the center of the image during the test phase. The crop image size for GoogLeNet~\cite{Szegedy_2015} and VGG-16 \cite{Simonyan_2014} models is $224 \times 224$, while for AlexNet \cite{Krizhevsky_2012} it is $227 \times 227$.

\subsection{Evaluation on the Multi-PIE Ear Dataset}
We have first assessed the performance of the deep CNN models on the collected Multi-PIE ear dataset. AlexNet \cite{Krizhevsky_2012}, VGG-16 \cite{Simonyan_2014}, and GoogLeNet \cite{Szegedy_2015} architectures have been employed and fine-tuned using their pretrained models that were trained on the ImageNet dataset \cite{Deng_2009}. The obtained results on the test set are listed in Table \ref{Multi-Pie}. In the table, the first column contains the name of the model, the second one contains the corresponding classification accuracy, and the third and fourth ones indicate whether augmentation and alignment have been applied or not. As can be seen, the achieved classification rates are quite high due to the controlled nature of the Multi-PIE ear dataset. VGG-16 model \cite{Simonyan_2014} is found to perform the best. Data augmentation has contributed around $1\%$ to the accuracy. Alignment did not lead to an improvement. However, this point requires further investigation, since no precise registration of the ear images has been done and they are only aligned roughly to one side.

\begin{figure*}[t]
	\centering{\includegraphics[scale=0.33]{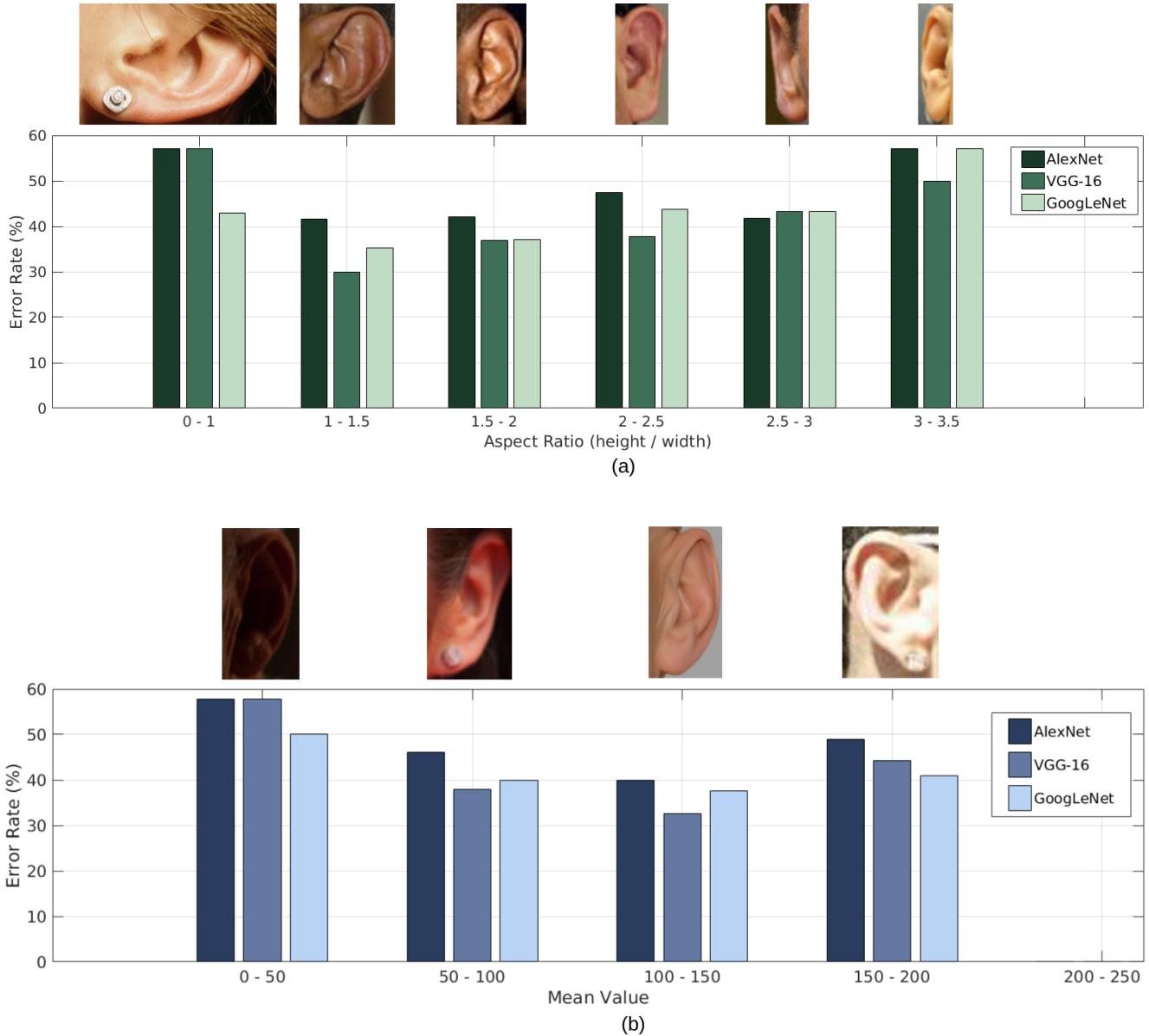}}
	\caption{UERC dataset test results (a) Sample ear images of different aspect ratios and the corresponding error rates for each aspect ratio interval, (b) Sample ear images of different average intensity values and the corresponding error rates for each average intensity interval.}
	\label{fig_error_rate}
\end{figure*}

\subsection{Evaluation on the UERC Dataset}
For the UERC dataset experiments, we have followed the experimental setup in \cite{Emersic_2017_c}. As in the experiments on the Multi-PIE ear dataset, AlexNet \cite{Krizhevsky_2012}, VGG-16 \cite{Simonyan_2014}, and GoogLeNet \cite{Szegedy_2015} architectures have been employed and fine-tuned using their pretrained models that were trained on the ImageNet dataset \cite{Deng_2009}. However, this time we have also applied a two-stage fine-tuning as described in Section \ref{subsec2.6}, that is we have first fine-tuned the pretrained deep CNN model on the Multi-PIE ear dataset and then fine-tuned the obtained updated model further on the training part of the UERC dataset. The experimental results are given in Table \ref{UERC_results}. In the table, the first column contains the name of the model, the second one contains the corresponding classification accuracy, the third one shows whether a single or two stage fine-tuning is applied, and the fourth and fifth ones indicate whether augmentation and alignment have been applied or not. For the third column, if the value is ImageNet, then in that experiment only one-stage fine-tuning has been performed and the pretrained model, which was trained on the ImageNet, has been fine-tuned using the training part of the UERC dataset. If the value is Multi-PIE, then two-stage fine-tuning has been applied, first on the Multi-PIE ear dataset, then on the training part of the UERC dataset.

Compared to the results in Table \ref{Multi-Pie}, the attained performance is significantly lower. Although the number of subjects to classify is less in the UERC dataset compared to the Multi-PIE ear dataset ---166 vs. 205---, due to challenging appearance variations and low quality images, ear recognition on the UERC dataset is a far more difficult problem.

The first three rows of the Table \ref{UERC_results} corresponds to the experimental results obtained in \cite{Emersic_2017_c}. For that study, the authors have employed AlexNet \cite{Krizhevsky_2012}, VGG-16 \cite{Simonyan_2014}, and SqueezeNet \cite{Iandola_2016}, and also utilized data augmentation. Comparing the accuracies obtained with AlexNet~\cite{Krizhevsky_2012} and VGG-16 \cite{Simonyan_2014} in \cite{Emersic_2017_c} and in our study under the same setup, that is with data augmentation and one-stage fine-tuning, it can be seen that our implementation has a slight improvement. In \cite{Emersic_2017_c}, 49.51\% and 51.25\% correct classification rates have been achieved using AlexNet~\cite{Krizhevsky_2012} and VGG-16 \cite{Simonyan_2014}, respectively, whereas in our study we have reached accuracies of 52\% and 54.2\%, respectively. This slight increase could be due to the differences in the parameters used for data augmentation and fine-tuning procedure.

From Table \ref{UERC_results}, it can be observed that the proposed two-stage fine-tuning procedure results in improved performance. For AlexNet \cite{Krizhevsky_2012}, with data augmentation and without alignment, the correct classification rate is increased from 52\% to 56.46\%. For VGG-16 \cite{Simonyan_2014} and GoogLeNet \cite{Szegedy_2015}, the increase is from 54.2\% to 63.62\% and from 55.02\% to 60.91\%, respectively. These significant improvements indicate that domain adaptation is indeed necessary and useful.  
This finding is in line with the results obtained in \cite{Ozbulak_2016}, where we have shown that when limited amount of training data is available for a task, it is more useful to transfer a pretrained model, which is trained on the images from the same domain. Specifically, for example, for age and gender classification, it is more useful to transfer a pretrained model, which is trained on face images, compared to transferring a pretrained model, which is trained on generic object images. In summary, compared to the results obtained with the VGG-16 \cite{Simonyan_2014} model in \cite{Emersic_2017_c}, we have achieved around 12\% absolute increase in performance ---51.25\% vs. 63.62\%. Similar to the results obtained on the Multi-PIE ear dataset, alignment did not lead to an improvement. Again, it should be noted that no precise registration of the ear images has been done and they are only aligned roughly to one side, therefore, this point requires further investigation. Among the employed models, VGG-16 model is found to be the best performing one.

We then fused the individual models in order to improve the performance further. For each model two-stage fine-tuning has been performed. Data augmentation has been applied and alignment has been omitted. We utilized the \textit{max rule} \cite{Kittler_1998} to combine the classification scores. We have employed five different confidence score calculation schemes ---basic, d2s, d2sr, avg-diff, diff1--- as listed in Table \ref{tab_fusion_method}. The results are given in Table \ref{UERC_Fusion}. The best performance is obtained when combining the best two performing models, that is VGG-16 \cite{Simonyan_2014} and GoogLeNet \cite{Szegedy_2015}, leading to 67.5\% correct classification, which is around 4\% higher than the one obtained with the single best performing model. No significant performance difference is observed between the employed confidence score calculation methods.

\begin{figure}[t]
	\centering{\includegraphics[scale=0.43]{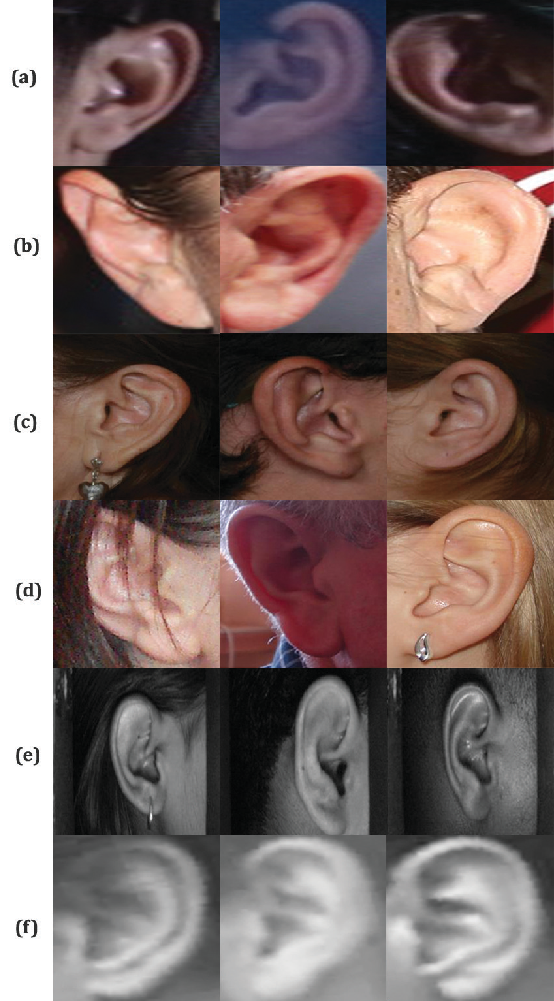}}
	\caption{Sample ear images from the datasets used for dataset identification experiments: (a) Multi-PIE Ear Dataset, (b) AWE, (c) AMI, (d) WPUT, (e) IITD, and (f) CP.}
	\label{fig_all_dataset}
\end{figure}

\begin{table*}[!h]
\renewcommand{\arraystretch}{1.3}
\caption{UERC dataset fusion results}
\label{UERC_Fusion}
\centering
\begin{tabular}{|c|c|c|c|c|c|}
\hline
Models  & Basic & d2s & d2sr & avg-diff & diff1\\
\hline
AlexNet + VGG-16 & 63.95\% & 64.06\% & 63.84\% & 63.95\% & 64.06\% \\
\hline
AlexNet + GoogLeNet & 63.51\% & 64.06\% & 64.16\% & 63.51\% & 63.73\% \\
\hline
VGG-16 + GoogLeNet & 67.53\% & 67.31\% & 67.53\% & 67.53\% & 67.42\% \\
\hline
All & 66.34\% & 66.01\% & 65.68\% & 66.34\% & 66.23\% \\
\hline
\end{tabular}
\end{table*}

\subsection{Effect of Image Quality on the Performance}
The effect of aspect ratio and illumination conditions of the image on the recognition performance has been analyzed. The results are shown in Fig. \ref{fig_error_rate}. As can be seen in Fig. \ref{fig_error_rate}(a), different aspect ratios occur due to varying view angles and ear shapes. Low aspect ratio, i.e. between 0-1, mainly implies in-plane rotated ear images, while higher aspect ratios, i.e. higher than 2, mainly refers to the cases of out-of-plane view variations. Experimental results show that the ear recognition system performs better, when the ear images are cropped from profile faces. Rotations of larger degrees and out-of-plane variations cause a performance drop. Samples of illumination variations from the UERC dataset can be seen in Fig. \ref{fig_error_rate}(b). Mean values in the x-axis correspond to the average intensities of the ear images. In the dark images, the details of the ear are not visible causing a loss of information. On the other hand, when the image is very bright, reflections and saturated intensity values are observed. Both of these conditions deteriorate the performance. 

\subsection{Dataset Identification}
During our ear recognition system development and training for the UERC challenge \cite{Emersic_2017_b}, we have tried to utilize the previously proposed ear datasets. We have combined them and used them for training. However, we could not have achieved a performance improvement. This outcome led us to consider the problem of dataset bias. In order to investigate this, we have designed an experiment, in which the class labels of the ear images are the names of the datasets that they belong to. That is, in this experiment, input to the deep CNN model is an ear image and the classification output is the name of the dataset that it belongs to. The goal was to observe whether the deep CNN model can distinguish the differences between the datasets. For this experiment six different ear datasets have been used, namely the Multi-PIE ear dataset, AWE~\cite{Emersic_2017_a}, AMI \cite{Gonzalez_2008}, WPUT \cite{Frejlichowski_2010}, IITD \cite{Kumar_2012}, and CP \cite{Carreira_Perpinan_1995} datasets. Sample images from these six datasets can be seen in Fig \ref{fig_all_dataset}. In this experiment, VGG-16 model \cite{Simonyan_2014} has been fine-tuned using the training parts of these datasets. Obtained training accuracy was 100\%. This fine-tuned model has achieved 99.71\% correct classification on the test set. Clearly, the system can easily identify ear images from different datasets. This is a very interesting and important outcome that requires further investigation in the future studies. 

\section{Conclusion}
In this study, we have addressed several aspects of ear recognition. First, we have proposed a two-stage fine-tuning strategy for deep convolutional neural networks in order to perform domain adaptation. For this approach, we have first constructed an ear dataset from the Multi-PIE face dataset \cite{Gross_2008_a,Gross_2008_b}, which we named as Multi-PIE ear dataset. In the first stage, we have fine-tuned the pretrained deep CNN models, which were trained on the ImageNet, on this newly collected dataset. This provides domain adaptation for the pretrained deep CNN models. In the second stage, we perform fine-tuning operation on the target dataset, which is the UERC dataset \cite{Emersic_2017_b}, in this work. This second stage provides a more specific domain and/or dataset adaptation. This step is also very crucial, since as we have shown in the experiments, there exists a dataset bias \cite{Torralba_2011} among the ear recognition datasets. We have also combined the deep CNN models to improve the performance further. Besides, we have analyzed in depth the effect of ear image quality, intensity level and aspect ratio, on the classification performance.

We have conducted extensive experiments on the UERC dataset~\cite{Emersic_2017_b}. We have shown that performing two-stage fine-tuning is very beneficial for ear recognition. With data augmentation and without alignment, for AlexNet \cite{Krizhevsky_2012}, the correct classification rate is increased from 52\% to 56.46\%. For VGG-16 \cite{Simonyan_2014} and GoogLeNet \cite{Szegedy_2015}, the increase is from 54.2\% to 63.62\% and from 55.02\% to 60.91\%, respectively. This consistent improvement indicates the importance of transferring a pretrained CNN model from a closer domain. It has been observed that combining different deep convolutional neural network models has led to further improvement in performance. We have achieved the best performance by combining the best two performing models, that is VGG-16 \cite{Simonyan_2014} and GoogLeNet \cite{Szegedy_2015}, leading to 67.5\% correct classification, which is around 4\% higher than the one obtained with the single best performing model. We have noticed that performing alignment did not improve the performance. However, this point requires further investigation, since the ear images have not been precisely registered and they have been only coarsely aligned by flipping them to one side. Effect of different aspect ratios, which have been resulted in due to varying view angles and ear shapes, and illumination conditions have also been studied. The ear recognition system performs better, when the ear images are cropped from profile faces. Very dark and very bright illumination causes missing details and reflections, which results in performance deterioration. Finally, we have conducted experiments to examine the dataset bias. Given an ear image as input, we were able to classify the dataset that it has come from with 99.71\% accuracy, which indicates a strong bias among the ear recognition datasets. For future work, we plan to address automatic ear detection, precise ear alignment, and dataset bias, which are important research problems in the ear recognition field.

\IEEEpubidadjcol

\ifCLASSOPTIONcompsoc
  \section*{Acknowledgments}
\else
  \section*{Acknowledgment}
\fi

This work was supported by the Istanbul Technical University Research Fund, ITU BAP, project no. 40893.

\end{document}